\documentclass[conference]{IEEEtran}
\IEEEoverridecommandlockouts
\usepackage{cite}
\usepackage{multirow}%
\usepackage{amsmath,amssymb,amsfonts}
\usepackage{xurl}
\usepackage[ruled,linesnumbered]{algorithm2e}
\SetAlgoNoLine
\DontPrintSemicolon
\SetKwInput{KwInput}{Input}
\SetKwInput{KwParameter}{Parameter}
\SetKwInput{KwOutput}{Output}
\usepackage{graphicx}
\usepackage{hyperref}
\usepackage{subcaption}
\usepackage{caption}
\usepackage[export]{adjustbox} % for valign=
\usepackage[numbers]{natbib}
\usepackage{tabularx}
\usepackage{booktabs}
\usepackage{textcomp}
\usepackage{xcolor}
\usepackage{enumitem}
\def\BibTeX{{\rm B\kern-.05em{\sc i\kern-.025em b}\kern-.08em
    T\kern-.1667em\lower.7ex\hbox{E}\kern-.125emX}}
\def\method{\texttt{M-LINKX}}
\usepackage{pifont}
\newcommand{\cnum}[1]{\ding{\numexpr171+#1\relax}}

\begin{document}
\bstctlcite{IEEEexample:BSTcontrol}
\title{M-LINKX: Multiview Graph Learning for Brain Cognitive Disease Detection
} %Neurocognitive Disorder Classification | EEG-Based Dementia Classification

 % \author{\IEEEauthorblockN{Anonymous Submission}
 % }

 \author{\IEEEauthorblockN{An Phan, Yufei Jin, and Xingquan Zhu}
\IEEEauthorblockA{\textit{Dept. of Electrical Engineering and Computer Science, Florida Atlantic University, Boca Raton, FL 33431, USA} \\
\{tphan2025, yjin2021, xzhu3\}@fau.edu}
 }
\maketitle

\begin{abstract}
Electroencephalogram (EEG) is a non-invasive and relatively low-cost procedure that measures brain electricity for the detection of cognitive diseases. EEG-based classification of dementia-related conditions, including Alzheimer's disease (AD), mild cognitive impairment (MCI), and frontotemporal dementia (FTD), remains challenging because EEG signals are noisy, non-stationary, and vary across subjects.
%This task remains challenging because EEG signals are noisy, non-stationary, and vary across subjects.
Segment-based learning provides a practical way to model long EEG recordings by converting them into fixed-length inputs. For each segment, discriminative information may be explored by using signals within each channel (\textit{i.e.} electrode), as well as interactions between EEG channels. % appear not only in channel-level signal characteristics, but also in frequency-specific interactions between EEG channels. 
In this paper, we propose \method{}, a multi-view graph learning framework for EEG-based dementia classification. For each segment, we extract channel-level node features and construct multiple functional-connectivity (FC) graph views, where each view is defined by a specific combination of connectivity metric, frequency band, and topology filter, respectively.
Instead of relying on message passing over the constructed graphs, \method{} follows a simple design in modeling node features and adjacency-based connectivity representations. The graph-view representations are fused using global trainable view weights, and subject-level prediction is obtained by averaging segment-level probabilities. 
Experiments on two three-class EEG datasets with different diagnostic groups, CAUEEG (HC/MCI/Dementia) and AHEAP (HC/AD/FTD), %CAUEEG and AHEAP, 
show that \method{} achieves the best subject-level performance under the main experimental settings. %, with balanced accuracy of 66.49\% and 61.76\%, respectively. 
Our study suggests that multi-view functional connectivity can improve EEG-based dementia classification when integrated with an appropriate graph-learning architecture. Code and data are available at \url{https://github.com/anphantt/MLINKX}.
\end{abstract}

% \hill{In the paper, we need to clearly answers several questions: (1) why using graphs to model electrode connectivity?; (2) why using LINKX type of models (which are used for heterophilic graphs); (3) why using multview to capture diverse connectivities.}

% \hill{Some thoughts in my mind. Brain congnitive disease is impacted by mailfunctional of brain regions. Using graphs can explicitly capture different brain regions' correlation for learning. However, brain functions are extremely complicated and signals such as EEG are very noisy. At any moment, our brain is rarely just working on one tasks, making it difficult to know when and how a regions is failling in respective tasks (e.g., walking, balancing, hearing, etc.) It's hard to capture genuine connection patterns between regsions. So instead of rely on one type of correlations, we rely on different types of connectivities. As for LINKX, the networks are essentially heterophilic (as research has shown such characteristics? \cite{}). Our experiments also clearly show that such method is better then complicated GCN or GAT models.}

% \hill{check following reference which concludes that "message-passing under both GNNs and Transformers can NOT be fully explored and exploited":
% https://icml.cc/virtual/2025/poster/45642}


\section{Introduction}

% %(1) motivation for dementia detection and EEG-based classification 
%Dementia is a group of progressive neurological disorders that impair cognition, behavior, communication, and daily functioning \cite{who_dementia_2025}. 
Dementia-related conditions, including Alzheimer's disease (AD), mild cognitive impairment (MCI), and frontotemporal dementia (FTD), create a growing global burden, with more than 55 million people currently living with dementia and projections reaching 139 million by 2050 \cite{alzint_dementia_statistics}. 
While AD accounts for approximately 60--70\% of dementia cases, %\cite{who_dementia_2025}, 
MCI may represent an early symptomatic stage before dementia \cite{salari_global_2025}, and FTD often affects younger individuals with behavioral, language, and communication impairments \cite{urso_incidence_2025}. 
These challenges motivate accessible and cost-effective tools for early detection and differential diagnosis. 
EEG is a promising modality because it is non-invasive, relatively low-cost, and more accessible than MRI, fMRI, or PET \cite{bi_eeg_2025}.

% (2) Related work of using Functional Connectivity (FC) of EEG in Alzheimer's and Dementia Detection: \\Beside using the raw EEG signal and put into LSTM, Transformer or 1D CNN, or use handcrafted features and ensemble, they also utilize the functional connectivity metrics which measure the interaction between channels. 
% - papers use EEG raw signal only (feed to CNN, LSTM), baseline 1D ResNet and CNN-LSTM \cite{li_mild_2023, miltiadous_dice-net_2023, sreedhar_fuzzy_2025, acharya_eegconvnext_2025, wang_study_2025, dharia_dual-transformer_2026}

Existing EEG-based dementia classification studies differ largely in how they represent the EEG signal. End-to-end deep learning approaches learn discriminative patterns directly from raw EEG using CNNs, LSTM-based, and transformer-based architectures \cite{xie_transformer-based_2022, sreedhar_fuzzy_2025, vo_extraction_2026}. %khosravi_fusing_2024
Other approaches first transform EEG signals into compact descriptors, such as relative band power, Hjorth parameters, wavelet energy, spectral entropy, and statistical features, and then use these representations with conventional machine learning, ensemble methods, or deep learning models \cite{ puri_alzheimers_2022, miltiadous_dice-net_2023, phan_edlad_2025}. These channel-wise representations can capture vital temporal or spectral abnormalities associated with dementia, but do not explicitly quantify how different EEG channels interact with each other. Since neurodegenerative changes may also affect communication among distributed brain regions, functional connectivity (FC) has been widely used as a complementary EEG representation in dementia studies. FC measures statistical dependency or synchronization between pairs of EEG channels, with common measures including coherence, phase-lag index (PLI), weighted phase-lag index (wPLI), and Pearson correlation \cite{jiang_classification_2025, zheng_time-frequency_2025, paitel_functional_2025}.

% \\How they use FC: - graph neural network - CNN for FC matrices. \\The biomarker from FC to classify the AD/MCI/FTD: strong/weak in [metric: PLI/Coherence...] in band [alpha, theta,...], which region show the disease signal stronger than other region ....\\

% Dementia-related changes may affect not only local EEG activity but also the communication between brain regions, FC provides a natural representation for connectivity-based models and graph-based learning approaches.

Recent EEG connectivity studies have shown that dementia-related FC alterations are not uniform across frequency bands, connectivity metrics, or brain regions. A recent systematic review reported that EEG connectivity studies in MCI and AD commonly use coherence, phase-locked measures, and graph-theoretic features; in addition to the frequently observed reduction of alpha-band connectivity, some studies also reported increased theta-band connectivity \cite{paitel_functional_2025}. Subject-level coherence from five frequency bands was represented as a connectivity tensor and used in a Coherence-CNN for AD, FTD, and cognitively normal classification \cite{jiang_classification_2025}. FC changes measured using Pearson correlation, mutual information, and PLI showed reduced theta-band connectivity in frontal regions, increased beta-band connectivity in posterior regions, and AD-specific reductions in central theta-band connectivity \cite{zheng_time-frequency_2025}.
Alpha-band FC was found to be most discriminative for separating dementia groups from healthy controls and delta-band FC contributed more to AD--FTD discrimination \cite{mlinaric_eeg-based_2026}. Together, these findings indicate that dementia-related EEG connectivity is heterogeneous and depends on how functional relationships are measured and represented. %These findings suggest that disease-related information may be distributed across multiple FC views rather than concentrated in a single connectivity representation. 

% Graph-based learning provides a natural framework for modeling EEG connectivity, where EEG channels are represented as nodes and functional connectivity (FC) values define edges through an adjacency matrix.
% \ann{(1) why using graphs to model electrode connectivity?}

% Such FC representations naturally lead to a graph-based formulation, where EEG channels are represented as nodes, channel-level EEG descriptors serve as node features, and FC values define edges through an adjacency matrix. In this formulation, the graph structure provides relational information among EEG channels. 
Because FC describes pairwise relationships between EEG channels, an FC matrix can be naturally formulated as a graph, where EEG channels are represented as nodes and FC values define edges through an adjacency matrix. In this formulation, channel-level EEG descriptors serve as node features, allowing the model to preserve local channel-specific information while the graph structure captures relational dependencies among electrode sites.
Recent EEG-based dementia studies have adopted graph models to capture such connectivity patterns \cite{wu_changes_2024, adebisi_eeg-based_2024, abadal_graph_2025, cao_identification_2025}. However, graph-based models are sensitive to how the adjacency matrix is constructed, because message passing is directly guided by the chosen graph topology. A~single connectivity graph may therefore provide only one approximation of the underlying neural interaction structure. In general graph learning, multi-view graph models address this limitation by constructing multiple graph views and learning complementary information across them. These views may correspond to different adjacency matrices, feature-topology channels, or relational assumptions, allowing the model to integrate diverse structural information rather than relying on one predefined graph \cite{wang2020gcn, yao_multi-view_2022, xiao_graph_2024, peng2025multi}.

% \ann{(3) why using multview to capture diverse connectivities?}

For EEG-based dementia classification, this multi-view idea can be translated into multiple FC graph representations of the same EEG segment. Instead of treating graph construction as a single fixed choice, different connectivity metrics, frequency bands, and topology filters can define complementary graph views that emphasize different channel-interaction patterns. Recent studies have begun to explore multi-view and frequency-specific graph representations for EEG-based dementia analysis. 
Multi-frequency graphs have been constructed using Pearson correlation and mutual information to capture both linear and nonlinear interactions \cite{liu_multi-frequency_2025}. Other work has combined functional and structural connectivity in a multi-graph convolutional framework with frequency-specific graph inputs \cite{xu_multi-graph_2025}. In addition, joint modeling of spatial, temporal, and spectral dependencies has been explored using a dual-path graph neural network \cite{zhang_dual_2025}. These studies suggest that dementia-related information may be distributed across different connectivity metrics, frequency bands, and graph structures, motivating a multi-view connectivity learning strategy rather than relying on a single predefined FC representation.

Motivated by this observation, we propose \method{}, a multi-view graph learning framework for EEG-based dementia classification. 
Inspired by LINKX~\cite{lim2021large}, a graph learning architecture originally designed for non-homophilous graph settings, \method{} separately models node features and adjacency information. 
This design is suitable for FC-based EEG graphs because edges represent statistical or phase-based dependencies between electrode signals rather than simple homophilic similarity between neighboring nodes; consistent with this view, recent brain network modeling work has questioned whether conventional message passing is always necessary for FC graphs, since FC matrices already encode pairwise relationships and message-passing models may reuse or smooth this information in a suboptimal way \cite{yang2025we}. 

% Contributions
\noindent This paper makes the following contributions.
\begin{itemize}[leftmargin=*, labelsep=0.5em, itemsep=0pt, topsep=2pt]
\item{We introduce \method{}, a multi-view graph learning framework for EEG-based dementia classification. In this framework, each EEG segment is represented using channel-level node features together with multiple FC graph views, where each view is defined by a specific combination of connectivity metric, frequency band, and graph topology.}
\item{We design a multi-view encoder that separately models node features and graph-view adjacency matrices representations. The graph-view representations are fused using global trainable view weights, allowing the model to assign different contributions to different connectivity views.}
\item{We evaluate \method{} on two dementia-related EEG datasets with different diagnostic groups and baselines. % compare it against raw-signal, feature-based, single-view graph, and multi-view graph baselines. 
Results show that \method{} achieves the best subject-level performance under the main experimental settings. %the proposed LINKX-bank model achieves stronger subject-level classification performance than using individual connectivity views or simpler fusion models.
}
% \item{We analyze the contribution of different connectivity views and show that dementia-related information is distributed across connectivity metrics, frequency bands, and graph topologies, supporting the need for a multi-view connectivity learning strategy.}
\end{itemize}
The remainder of this paper is organized as follows. 
Section II defines the notation and problem setting.  % Section II defines the notation and formulates the EEG-based dementia classification problem. 
Section III presents the proposed \method{} methodology. % Section III describes the proposed \method{} framework, including EEG feature extraction, functional connectivity construction, graph view generation, and model architecture.
Section IV reports the experimental results and analysis.  % Section IV describes the experimental setup, baseline comparisons, ablation studies, and classification results.
Section V concludes the paper and discusses future work.  %Section V concludes the paper and discusses future research directions.

\section{Notation and Problem Definition}\label{sec:notation}
Let $\mathcal{C} = \{c_1,\ldots,c_C\}$ denote the set of EEG channels, where $C = |\mathcal{C}|$ is the number of channels. Each subject $\mathbf{S}_i$ has $N_i$ EEG segment represented as
\[
{s}_{i,t}\in\mathbb{R}^{C\times T},
\]
where $i$ indexes the subject, $t$ indexes the segment from that subject, and $T$ is the number of time points in each segment.

Given a set of labeled training subjects,
\begin{equation}
\mathcal{D}{\mathrm{train}} = \{(\mathbf{S}_i,y_i)\}_{i=1}^{D}, \hspace{5pt} y_i \in {0,\ldots,L-1},
\end{equation}
The learning \textbf{objective} is to train a classification model from $\mathcal{D}{\mathrm{train}}$ and classify previously unseen test subjects with the highest performance measures. 

%{\textit{Graph Representation of EEG Signals}}
\subsection{Graph Representation of EEG Signals}
% \hill{why segment?}
% \ann{TO DO}

EEG recordings may vary in duration across subjects and are often characterized by non-stationary temporal patterns. We divide each recording into fixed-length overlapping segments using a sliding-window strategy. This segment-based representation provides consistent input dimensions, makes long EEG recordings more computationally manageable, and supports the subsequent feature extraction step.

As shown in the upper-left part of Fig.~\ref{fig:framework}, each EEG recording is first segmented, and each segment is then used for feature extraction and graph construction in the proposed framework. 
%ref the fig 1 
For each segment $s_{i,t}$%$\mathcal{S}_{i,t}$
, we extract channel-level node features and form a node attribute matrix
\[
\mathbf{X}_{i,t}\in\mathbb{R}^{C\times F},
\]
where $F$ is the number of node features. Each row of $\mathbf{X}_{i,t}$ corresponds to one EEG channel.
We construct $K$ functional-connectivity graph views for each segment. The $k$-th graph view is represented by an adjacency matrix
\[
\mathbf{A}^{(k)}_{i,t}\in\mathbb{R}^{C\times C},
\]
% where each view may correspond to a specific combination of connectivity metric, frequency band, and topology filter. 
The set of graph views for segment  $s_{i,t}$ is denoted as
\[
\mathcal{A}_{i,t}=\{\mathbf{A}^{(1)}_{i,t},\mathbf{A}^{(2)}_{i,t},\ldots,\mathbf{A}^{(K)}_{i,t}\}.
\]
Thus, each EEG segment is represented by
\[
\mathcal{G}_{i,t}=\left(\mathbf{X}_{i,t},\mathcal{A}_{i,t}\right),
\]
where $\mathbf{X}_{i,t}$ contains channel-level node attributes and $\mathcal{A}_{i,t}$ contains multiple connectivity-based graph views.

\section{Methodology}

\subsection{Overall Framework}
\begin{figure*}[t]
    \centering
    \includegraphics[width=\linewidth]{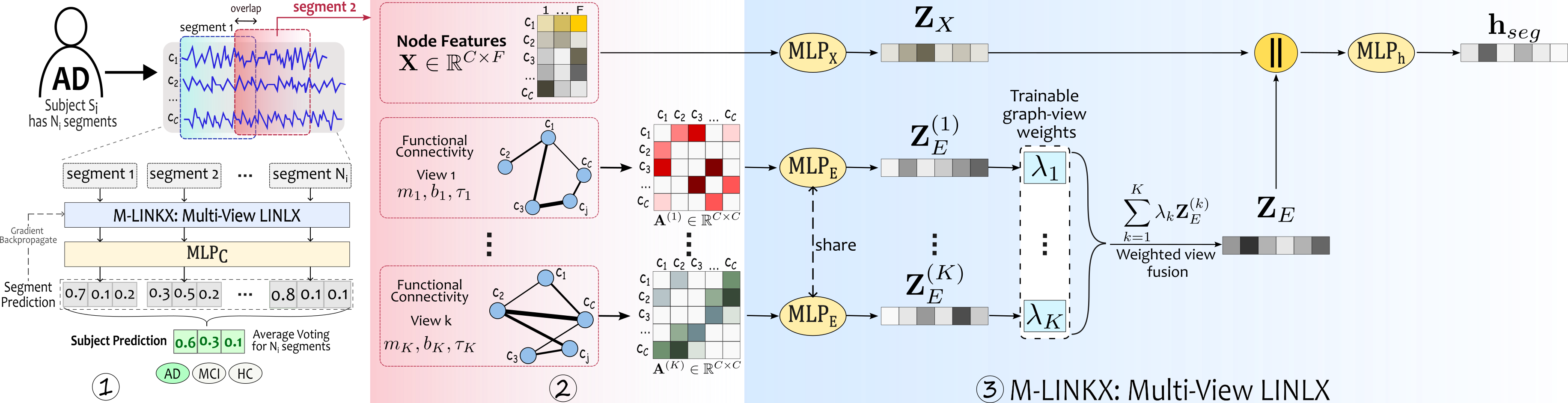}
\caption{Overview of the proposed \method{} framework. For readability, block \cnum{1} should be read from top to bottom, whereas blocks \cnum{2} and \cnum{3} should be read from left to right. Block \cnum{1} shows the subject-level classification pipeline: each EEG recording is divided into fixed-length overlapping segments, segment-level probabilities are predicted, and subject-level prediction is obtained by averaging probabilities across segments. Block \cnum{2} shows feature extraction and graph construction for a segment, where ``segment 2'' is represented by channel-level node features and multiple FC graph views from different connectivity metrics, frequency bands, and topology filters. Block \cnum{3} shows the \method{} encoder, which learns node-feature and graph-view embeddings, fuses graph-view embeddings using global trainable view weights, and produces segment-level predictions.
}

    % \caption{Overview of the proposed \method{} framework. For readability, block \cnum{1} should be read from top to bottom, whereas blocks \cnum{2} and \cnum{3} should be read from left to right. The overall subject-level classification pipeline is summarizes in \cnum{1}. Each subject-level EEG recording is divided into fixed-length overlapping segments. Each segment is then used for feature extraction and graph construction in \cnum{2}, then processed by \method{} in \cnum{3} to produce segment-level predictions. The final subject-level prediction is obtained by averaging the segment-level probabilities across all segments from the same subject. \cnum{2} details how each EEG segment represented by channel-level node features and multiple FC graph views generated from different combinations of connectivity metrics, frequency bands, and topology filters. The \method{} encoder in \cnum{3} learns a node-feature embedding and separate graph-view embeddings, then combines the graph-view embeddings using global trainable view weights shared across all segments and subjects. The fused segment representation is passed to a classifier to produce segment-level predictions. 
    % }
    \label{fig:framework}
    \vspace{-0.8em}
\end{figure*}

% \subsection{Proposed model}
Fig.~\ref{fig:framework} illustrates how the \method{} framework works. Each subject-level EEG recording is divided into fixed-length overlapping segments. Since diagnostic labels are available only at the subject level, each segment inherits the label of its corresponding subject during training.  Our goal is to learn an encoder that maps each segment-level graph input $\mathcal{G}_{i,t}$ to a segment embedding $\mathbf{h}_{i,t}$. A classifier then maps $\mathbf{h}_{i,t}$ to a class-probability vector $\mathbf{p}_{i,t}$. 
For a previously unseen subject, subject-level prediction is obtained by averaging segment-level probabilities:
\begin{equation}
\bar{\mathbf{p}}_i = \frac{1}{N_i}\sum_{t=1}^{N_i}\mathbf{p}_{i,t},
\end{equation}
and the predicted label is $\hat{y}_i=\operatorname{argmax}\bar{\mathbf{p}}_{i}$, where $\operatorname{argmax}(\bar{\mathbf{p}}_i)$ returns the class index with the highest averaged probability.

Throughout the following sections, we describe the model for a generic EEG segment and omit the subject and segment indices $(i,t)$ when there is no ambiguity. Thus, $\mathbf{X}$ denotes the node feature matrix of a segment, $\mathbf{A}^{(k)}$ denotes its $k$-th graph-view adjacency matrix, and $\mathcal{A}=\{\mathbf{A}^{(1)},\ldots,\mathbf{A}^{(K)}\}$ denotes its graph-view bank.

For each segment, channel-wise node features and functional-connectivity-based graph views are extracted as model inputs. The \method{} encoder integrates node-feature and graph-view representations to produce a segment embedding, which is used for segment-level classification. During inference, subject-level prediction is obtained by soft voting the predicted probabilities across all segments.

\subsection{Node Feature Extraction}
For each EEG segment, %\mathcal{S}_{i,t}$
channel-wise features are extracted to describe the local signal characteristics of individual electrodes. These features are concatenated within each EEG channel to form the node feature matrix
\begin{equation}
    \mathbf{X}
    =
    [\mathbf{x}_{1}, \mathbf{x}_{2}, \ldots, \mathbf{x}_{C}]^\top
    \in \mathbb{R}^{C \times F},
\end{equation}
where $\mathbf{x}_{j}\in\mathbb{R}^{F}$ denotes the feature vector of the $j$-th EEG channel. Each row of $\mathbf{X}$ corresponds to one EEG channel, and each column corresponds to one selected node feature.
%$\mathbf{X}$ defined in Section~\ref{sec:notation}. Each row of $\mathbf{X}$ corresponds to one EEG channel, and each column corresponds to one selected node feature.
% We use two %three
% groups of node features: relative band power (RBP), 
% %Hjorth parameters, 
% and statistical descriptors. The RBP features are computed from the power spectral density of each EEG channel using Welch's method~\cite{welch1967use}. 
% We extract five frequency-band features: delta $\delta$ ($1$--$4$ Hz), theta $\theta$ ($4$--$8$ Hz), alpha $\alpha$ ($8$--$13$ Hz), beta $\beta$ ($13$--$30$ Hz), and gamma $\gamma$ ($30$--$45$ Hz).
% % The Hjorth features include activity, mobility, and complexity, which provide compact time-domain measures of signal variance and temporal dynamics~\cite{hjorth1970eeg}. 
% The statistical descriptors summarize the amplitude distribution of each channel and include mean, standard deviation, skewness, kurtosis, minimum, maximum, and peak-to-peak amplitude.

% The selected features are concatenated for each EEG channel to form a node feature vector, the resulting node feature matrix is
% \begin{equation}
%     \mathbf{X}
%     =
%     [\mathbf{x}_{1}, \mathbf{x}_{2}, \ldots, \mathbf{x}_{C}]^\top
%     \in \mathbb{R}^{C \times F},
% \end{equation}
% where $\mathbf{x}_{j}\in\mathbb{R}^{F}$ denotes the feature vector of the $j$-th EEG channel. %and $F$ is the total number of selected node features.

\subsection{Functional Connectivity Extraction}

Given a preprocessed EEG segment, %\mathcal{S}_{i,t}$
functional connectivity (FC) is computed between pairs of EEG channels. %where $C$ is the number of EEG channels and $T$ is the number of time points, 
For a connectivity metric $m$ and frequency band $b$, this produces a dense FC matrix %$\mathbf{W_{i,t}^{m,b} \in \mathbb{R}^{C \times C}}$, 
$\mathbf{W}^{(m,b)} \in \mathbb{R}^{C \times C}$, 
where $w_{pq}^{(m,b)}$ denotes the connectivity strength between channels $c_p$ and $c_q$.
% each FC metric computes an adjacency matrix for each frequency band $b$: $A^{(b)} \in \mathbb{R}^{C \times C}$. The entry $A_{ij}^{(b)}$ denotes the connectivity strength between EEG channels $i$ and $j$ in band $b$.

In this work, we use coherence and the weighted phase lag index (wPLI) as connectivity metrics to construct band-specific FC matrices. %For each EEG segment, the output shape of each metric is $B \times C \times C$, where $B$ is the number of frequency bands. 

\textbf{Magnitude-Squared Coherence (\textit{Coh}).}
Coherence is a frequency-domain measure of linear coupling between two EEG signals. It measures how strongly two channels share consistent oscillatory activity at frequency $f$ \cite{nunez2006electric}. 
For channels $c_p$ and $c_q$, coherence is computed from the cross-spectral density $G_{pq}(f)$ and the auto-spectral densities $G_{pp}(f)$ and $G_{qq}(f)$ at frequency $f$ \cite{bendat2011random}:
\begin{equation}
    C_{pq}(f) = \frac{|G_{pq}(f)|^2}{G_{pp}(f)G_{qq}(f)}.
\end{equation}
The band-level coherence value is then calculated as the mean coherence over frequencies within band $b$:
\begin{equation}
    w_{pq}^{(\mathrm{Coh},b)}
    =
    \frac{1}{|\mathcal{F}_b|}
    \sum_{f \in \mathcal{F}_b}
    C_{pq}(f).
\end{equation}
Coherence value lies in $[0,1]$, where $0$ indicates no linear coupling and $1$ indicates perfect linear coupling.

\textbf{Weighted Phase Lag Index (\textit{wPLI}).}
wPLI is a phase-based connectivity metric that measures whether two EEG channels have a consistent non-zero phase-lag relationship\cite{vinck_improved_2011}. 

For each frequency band $b$, the signal from channel $c_p$ is first band-pass filtered, yielding $r_p^{(b)}[\ell]$, where $\ell=1,\ldots,T$ indexes time samples. The analytic signal is then obtained using the Hilbert transform:
\begin{equation*}
    z_p^{(b)}[\ell]
    =
    r_p^{(b)}[\ell]
    +
    j\mathcal{H}\left\{r_p^{(b)}[\ell]\right\},
\end{equation*}
where $\mathcal{H}\{\cdot\}$ denotes the Hilbert transform and $j$ is the imaginary unit \cite{gabor_theory_1947}. The instantaneous phase is extracted as
\begin{equation*}
    \phi_p^{(b)}[\ell]
    =
    \arg\left(z_p^{(b)}[\ell]\right).
\end{equation*}

For channels $c_p$ and $c_q$, the phase difference is defined as $\Delta\phi_{pq}^{(b)}[\ell] = \phi_p^{(b)}[\ell] - \phi_q^{(b)}[\ell]$. In our implementation, wPLI is estimated from the sine of the phase difference:
\begin{equation}
    w_{pq}^{(\mathrm{wPLI},b)}
    =
    \frac{
    \left|
    \sum_{\ell=1}^{T}
    \sin\left(\Delta\phi_{pq}^{(b)}[\ell]\right)
    \right|
    }{
    \sum_{\ell=1}^{T}
    \left|
    \sin\left(\Delta\phi_{pq}^{(b)}[\ell]\right)
    \right|
    }.
\end{equation}

wPLI value ranges from 0 to 1, with larger values indicating stronger and more consistent non-zero phase-lag coupling. 
%where values close to $0$ indicate weak or inconsistent non-zero phase-lag coupling, and values close to $1$ indicate strong and consistent non-zero phase-lag coupling. 

The resulting connectivity matrices are symmetric, with diagonal entries set to zero to exclude self-connections. Each selected metric-band pair is treated as one connectivity view and converted into a graph topology for multi-view learning.

\subsection{Graph Construction}
For each EEG segment, we construct a graph representation with each channel being a node and the FC values defining weighted edges between those nodes. 

% In the proposed framework, each graph view is defined by three components: the connectivity metric $m$, the frequency band $b$, and the topology filter $\tau$. The connectivity metric and frequency band first produce a dense weighted FC matrix $\mathbf{W}^{(m,b)}\in\mathbb{R}^{C\times C}$. 
Given the dense FC matrix $\mathbf{W}^{(m,b)}$ defined above, each graph view is specified by a connectivity metric $m$, frequency band $b$, and topology filter $\tau$. The topology filter then determines which edges are retained. 

Given a topology mask $\mathbf{M}^{(m,b,\tau)}\in\{0,1\}^{C\times C}$, the final adjacency matrix is obtained as
\begin{equation}
\mathbf{A}^{(k)}=\mathbf{A}^{(m,b,\tau)}
=\mathbf{W}^{(m,b)} \odot \mathbf{M}^{(m,b,\tau)},
\end{equation}
where 
$\odot$ denotes element-wise multiplication. 

%The connectivity metric and frequency band determine the dense FC matrix, while the topology filter determines which edges are retained. Given a dense FC matrix $A_{i,t}^{(m,b)}$, we apply a topology filter $\tau$ to obtain the final adjacency matrix of the graph:
%$M_{i,t}^{(m,b,\tau)} \in \{0,1\}^{C \times C}$ is a topology mask and 
%Since each subject contains multiple EEG segments, subject $s$ is represented by a set of segment-level graph inputs. The graph is represented as
% $\mathcal{G}_{s,r} = (X_{s,r}, A_{s,r})$, where $X_{s,r} \in \mathbb{R}^{C \times F}$ is the node feature matrix of the
% $r$-th segment from subject $s$, $C$ is the number of EEG channels, and $F$ is the number of node features. The adjacency matrix $A_{s,r} \in \mathbb{R}^{C \times C}$ represents the FC between EEG channels. 

% Therefore, each graph view is represented as
% \begin{equation}
%     \mathcal{G}_{s,r}^{(m,b,\tau)}
%     =
%     \left(
%     X_{s,r},
%     A_{s,r}^{(m,b,\tau)}
%     \right).
% \end{equation}

%  The \texttt{\textbf{full}} topology keeps all pairwise channel connections, whereas sparse topologies retain only selected edges. For example, \texttt{\textbf{fixed}}
% topology preserves predefined channel relationships, \texttt{\textbf{top-$k$}}  topology keeps the strongest connectivity edges, and \texttt{\textbf{combined}}   topology integrates predefined
% and data-driven edge selection. 
We consider several topology filters, including \textit{complete}, \textit{domain}, top-$k$, and \textit{hybrid}. 
\textbf{\textit{Complete topology}} keeps all pairwise channel connections, which forms a complete graph with edge weight denoting FC values between electrodes. \textbf{\textit{Domain topology}} keeps a predefined sparse graph based on spatially neighboring electrodes in the standard 10--20 EEG montage. \textbf{\textit{Top-$k$ topology}} keeps the strongest connectivity edges for each node. \textbf{\textit{Hybrid topology}} combines the domain topology with top-$k$ edges.
Thus, \textit{complete} and \textit{domain} have fixed edge sets across segments, whereas top-$k$ and \textit{hybrid} can vary in both retained edges and edge weights.
% \vspace{0.1cm}\noindent\textbf{\textit{Complete topology} keeps all pairwise channel connections, which forms a complete graph with edge weight denoting FC values between electrodes. 

% \vspace{0.1cm}\noindent\textbf{\textit{Domain topology} keeps a predefined sparse graph based on spatially neighboring electrodes in the standard 10--20 EEG montage. 
% Because it is independent of the subject, segment, class label, and connectivity values, it provides a stable spatial-neighborhood prior and reduces the redundancy of dense connectivity graphs. 

% \vspace{0.1cm}\noindent\textbf{\textit{Top-$k$ topology} keeps the strongest connectivity edges for each node. %Both \textit{complete} and \textit{domain} have fixed edge sets across subjects and segments; only their edge weights vary according to the connectivity values of each EEG segment. 
%In contrast, {top-$k$} topology keeps the strongest connectivity edges for each node.

% \vspace{0.1cm}\noindent \textbf{\textit{Hybrid topology}} combines the domain topology with top-$k$ edges.%to form \textit{hybrid} topology. 
% Thus, \textit{complete} and \textit{domain} have fixed edge sets across segments, whereas top-$k$ and \textit{hybrid} can vary in both retained edges and edge weights.
% Thus, top-$k$ and \textit{hybrid} are adaptive filters, where both the retained edges and their weights can change across segments. 

\begin{figure}
    \centering
    \includegraphics[width=\linewidth]{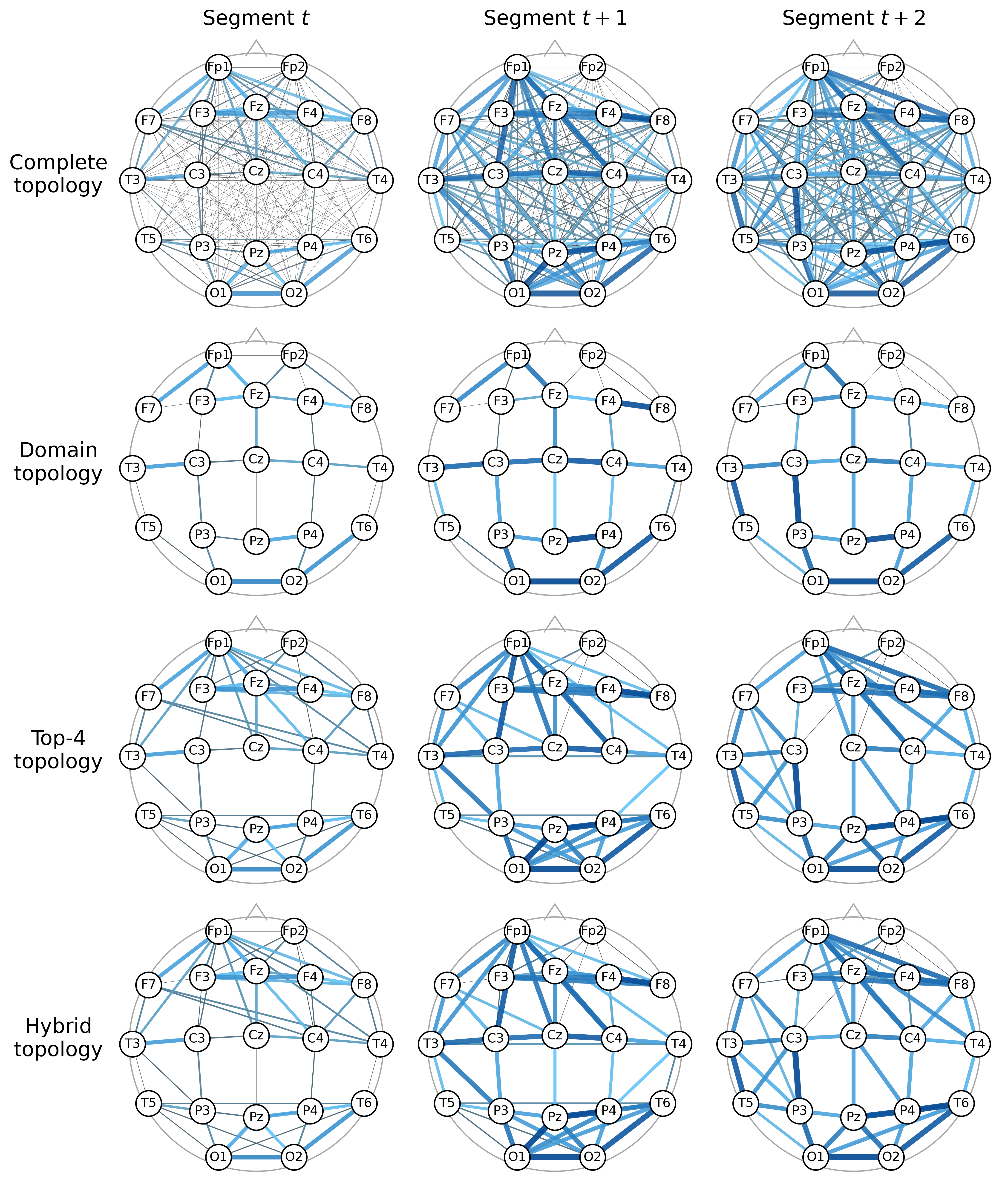}
    \caption{Topology-specific graph representations of alpha-band coherence from three consecutive EEG segments of the same subject recording. 
Nodes denote EEG channels arranged by approximate 10--20 scalp locations, and weighted edges denote retained coherence connections. Rows show the four topology filters: \textit{complete}, \textit{domain}, top-$k$, and \textit{hybrid}; columns show consecutive EEG segments. Edge thickness and color indicate coherence weight, with thicker and darker blue edges representing larger weights and thinner gray edges representing smaller weights. %For \textit{complete} and \textit{domain}, the edge set is fixed and only the weights vary over time, whereas top-$k$ and \textit{hybrid} can vary in both edge weights and retained topology. %\ann{update figure!}
% The panels compare \texttt{full} weighted functional connectivity, \texttt{fixed} predefined connections, top-$k$ node-wise top-4 strongest-neighbor connectivity, and \texttt{combined} topology using both predefined and strongest connections.
}
    \label{fig:topology_views}
\end{figure}

Fig.~\ref{fig:topology_views} illustrates these topology filters using alpha-band coherence, \textit{i.e.}, $m=\mathrm{coh}$ and $b=\alpha$, across consecutive EEG segments from the same recording. The figure shows that segment-level FC patterns can change across adjacent segments, with some segments showing noticeable topology changes and others preserving similar dominant connections.
% Fig.~\ref{fig:topology_views} illustrates these topology filters by fixing the graph view to alpha-band coherence, i.e. $m={coh}$ and $b=\alpha$, and varying the topology filter $\tau$ across consecutive EEG segments from the same recording. This example highlights the dynamic nature of segment-level EEG functional connectivity. Even when the same connectivity metric and frequency band are used, the connectivity pattern can vary substantially across adjacent segments. In some cases, the strongest connections change noticeably from one segment to the next, whereas other consecutive segments preserve similar dominant connections with only minor changes in edge strength. 

% Thus, for each segment $\mathcal{S}_{i,t}$, the multi-view graph representation shares the same node feature matrix $\mathbf{X}_{i,t}$ but uses different adjacency matrices generated from different connectivity metrics, frequency bands, and topology filters.

%These topology filters define how edges are selected from the dense functional-connectivity matrix. An example of these topology-specific graph views is shown in Fig.~\ref{fig:topology_views}.

% \ann{Fig.... illustrates how different topology filters convert the same dense connectivity matrix into different graph structures.}

% \subsection{Segment-Level Training and Subject-Level Prediction}

\subsection{M-LINKX Algorithm}
% Algorithm~\ref{alg:mlinkx} outlines major steps of the proposed \method. Given a train set with $N_{train}$ subjects, and $K$ graph views that define functional connectivity between EEG channels...\hill{briefly outline major steps} 
Algorithm~\ref{alg:mlinkx} outlines the major steps of the proposed \method. Given a training set with $D$ subjects and $K$ graph-view definitions, the \textsc{GenerateMultiViewGraphs} function follows the graph construction process and converts each EEG segment into a graph, with the subject label being propagated to all segments of the subject for learning. For each test subject, \method\ predicts probabilities of all segments of a test subject. The probabilities are averaged to obtain the subject-level prediction $\hat{y}_{test}$.

For each segment, \method{} follows a LINKX-style design to encode node features and graph-view adjacency matrices separately. The graph-view embeddings are combined using trainable view weights, then concatenated with the node-feature embedding and passed through an MLP for classification. Unlike the original LINKX, we omit additive skip connections between adjacency and node-feature embeddings.

% In this work, we use a LINKX-style encoder in the sense that node features an,, the function \textsc{GenerateMultiViewGraphs} follows the graph construction process described above and converts each EEG recording into segment-level graph samplesadjacency information are encoded separately. The original LINKX formulation combines the adjacency embedding and node-feature embedding using additive skip connections. In our implementation, we do not use these skip connections; instead, the node-feature embedding and the fused connectivity embedding are concatenated and passed through an MLP to obtain the final segment-level representation.

\begin{algorithm}[t]
\footnotesize
\caption{M-LINKX Algorithm}
\label{alg:mlinkx}
\KwInput{%
\begin{tabular}[t]{@{}l@{}}
$\mathcal{D}_{\mathrm{train}}=\{(\mathbf{S}_i,y_i)\}_{i=1}^{D};
\quad \mathbf{S}_{\mathrm{test}};$ \\[2pt]
graph-view definitions  $\mathcal{V}=\{(m_k,b_k,\tau_k)\}_{k=1}^{K}.$
\end{tabular}
}
\vspace{2pt}
% \KwInput{$\mathcal{D}_{\mathrm{train}}=\{(\mathbf{S}_i,y_i)\}_{i=1}^{D}$; $\mathbf{S}_{\mathrm{test}}$; graph-view definitions $\mathcal{V}=\{(m_k,b_k,\tau_k)\}_{k=1}^{K}$}
\KwParameter{\texttt{epochs}; $\Theta$; trainable view logits $\boldsymbol{\psi}\in\mathbb{R}^{K}$}
\vspace{2pt}
\KwOutput{$\hat{y}_{\mathrm{test}}$}

\SetKwFunction{GenGraphs}{GenerateMultiViewGraphs}

% \SetKwProg{Fn}{Function}{:}{end}

% \Fn{\GenGraphs{$S_i,\mathcal{V}$}}{
%     $\{\mathbf{S}_{i,t}\}_{t=1}^{N_i}\leftarrow \mathrm{Segment}(S_i)$\;
%     \For{$t=1$ \KwTo $N_i$}{
%         $\mathbf{X}_{i,t}\leftarrow f_{\mathrm{feat}}(\mathbf{S}_{i,t})$\;
%         \For{$k=1$ \KwTo $K$}{
%             $\mathbf{A}^{(k)}_{i,t}\leftarrow f_{\mathrm{graph}}(\mathbf{S}_{i,t};m_k,b_k,\tau_k)$\;
%         }
%         $\mathcal{A}_{i,t}\leftarrow\{\mathbf{A}^{(k)}_{i,t}\}_{k=1}^{K}$\;
%         $\mathcal{G}_{i,t}\leftarrow(\mathbf{X}_{i,t},\mathcal{A}_{i,t})$\;
%     }
%     \Return $\{\mathcal{G}_{i,t}\}_{t=1}^{N_i}$\;
% }
\vspace{2pt}
$\mathcal{D}^{\mathrm{seg}}_{\mathrm{train}}\leftarrow \emptyset$\;
\vspace{2pt}
\ForEach{$(\mathbf{S}_i,y_i)\in\mathcal{D}_{\mathrm{train}}$}{
    $\{\mathcal{G}_{i,t}\}_{t=1}^{N_i}\leftarrow \GenGraphs(\mathbf{S}_i,\mathcal{V})$\;
    \vspace{1pt}
    $\mathcal{D}^{\mathrm{seg}}_{\mathrm{train}}\leftarrow
    \mathcal{D}^{\mathrm{seg}}_{\mathrm{train}}\cup
    \{(\mathcal{G}_{i,t},y_i)\}_{t=1}^{N_i}$\;
}
\vspace{2pt}
\For{$e=1$ \KwTo \texttt{epochs}}{
        \vspace{1pt}    \ForEach{$(\mathcal{G}_{i,t},y_i)\in\mathcal{D}^{\mathrm{seg}}_{\mathrm{train}}$}{
        \vspace{1pt}
        $\mathbf{Z}^{X}_{i,t}\leftarrow f_{\mathrm{node}}(\mathbf{X}_{i,t})$\;
        \vspace{1pt}
        $\mathbf{Z}^{(k)}_{i,t}\leftarrow f_{\mathrm{adj}}(\mathbf{A}^{(k)}_{i,t}),\quad k=1,\ldots,K$\;
        \vspace{1pt}
        $\boldsymbol{\lambda}\leftarrow \operatorname{softmax}(\boldsymbol{\psi})$\;
        \vspace{1pt}
        $\mathbf{Z}^{E}_{i,t}\leftarrow \sum_{k=1}^{K}\lambda_k\mathbf{Z}^{(k)}_{i,t}$\;
        \vspace{1pt}
        $\mathbf{h}_{i,t}\leftarrow f_{\mathrm{fuse}}(\mathbf{Z}^{X}_{i,t}\,\|\,\mathbf{Z}^{E}_{i,t})$\;
        \vspace{1pt}
        $\mathbf{o}_{i,t}\leftarrow f_{\mathrm{cls}}(\mathbf{h}_{i,t})$\;
        \vspace{1pt}
        $\mathcal{L}\leftarrow \operatorname{CE}(\mathbf{o}_{i,t},y_i)$\;
        \vspace{1pt}
        $\Theta,\boldsymbol{\psi}\leftarrow \operatorname{Update}(\Theta,\boldsymbol{\psi};\mathcal{L})$\;
    }
}

$\{\mathcal{G}_{\mathrm{test},t}\}_{t=1}^{N_{\mathrm{test}}}
\leftarrow \GenGraphs(\mathbf{S}_{\mathrm{test}},\mathcal{V})$\;
\vspace{3pt}
$\boldsymbol{\lambda}\leftarrow \operatorname{softmax}(\boldsymbol{\psi})$\;
\vspace{2pt}
$\mathbf{p}_{\mathrm{test},t}\leftarrow
\mathrm{M\mbox{-}LINKX}(\mathcal{G}_{\mathrm{test},t};\Theta,\boldsymbol{\lambda}),
\quad t=1,\ldots,N_{\mathrm{test}}$\;
\vspace{2pt}
$\hat{y}_{\mathrm{test}}\leftarrow
\operatorname{argmax}\left(
\frac{1}{N_{\mathrm{test}}}
\sum_{t=1}^{N_{\mathrm{test}}}
\mathbf{p}_{\mathrm{test},t}
\right)$\;

\Return $\hat{y}_{\mathrm{test}}$\;

\end{algorithm}

\section{Experiments and Results}

\subsection{Datasets}
% \ann{AHEAP, CAUEEG information about number of subjects and class distribution, channels, sampling rate, eye state, duration of recording}

We evaluate the proposed approach on two EEG datasets for dementia-related classification. AHEAP (\textit{ds004504}) contains resting-state EEG recordings from 88 subjects across Alzheimer's disease (AD), frontotemporal dementia (FTD), and healthy control (HC) groups~\cite{ntetska_complementary_nodate}. CAUEEG contains EEG recordings from 1122 subjects across dementia, mild cognitive impairment (MCI), and healthy control (HC) classes~\cite{kim_deep_2023}. For CAUEEG, we use the non-overlap version, where validation and test subjects do not overlap with the training set. Dataset characteristics are summarized in Table~\ref{tab:desc}.

\begin{table}[t]
\centering
\caption{Summary of the EEG datasets used in this study.}
\label{tab:desc}

\footnotesize
\setlength{\tabcolsep}{3pt}
\resizebox{\linewidth}{!}{%
\begin{tabular}{@{}lll@{}}
\toprule
\textbf{Property} & \textbf{AHEAP} & \textbf{CAUEEG} \\ 
\midrule
\# of Subjects
& 36 AD, 23 FTD, 29 HC
& 291 Dementia, 395 MCI, 436 HC \\

\# of Channels
& 19 
& 19 \\

Condition 
& Eyes closed 
& Eyes closed and eyes open \\

Sampling Rate 
& 500 Hz 
& 200 Hz \\

Avg. Duration 
& 13.21 $\pm$ 2.32 min
& 12.80 $\pm$ 2.65 min \\
\bottomrule
\end{tabular}%
}
\end{table}

We use the EEG recordings provided by the original dataset sources without additional manual cleaning, artifact rejection, or channel interpolation. For AHEAP, we use the released preprocessed signals, which were band-pass filtered between 0.5 and 45 Hz and cleaned using independent component analysis (ICA). For CAUEEG, we use the released EEG recordings as provided and remove the two non-EEG auxiliary channels corresponding to EKG and photic stimulation.

For both datasets, we retain the common 19 EEG channels: Fp1, Fp2, F7, F3, Fz, F4, F8, T3, C3, Cz, C4, T4, T5, P3, Pz, P4, T6, O1, and O2. EEG recordings are divided into overlapping segments using a sliding window with 50\% overlap. The resulting segments are used to extract node features, construct functional-connectivity matrices, and form graph instances for model training. The majority of experiments reported in the paper are based on a 4$s$ segment for AHEAP and a 10$s$ segment for CAUEEG (because the former has 2.5 times higher sampling frequency). In addition, we also report results \textit{w.r.t.} different segment lengths for analysis.
% \subsection{Functional Connectivity Visualization}

% \ann{Some images showing that the connectivity matrices are different between classes and frequency band in both datasets}

% To further examine the motivation for using multiple connectivity views, we visualize class-wise functional connectivity patterns across different frequency bands and datasets. For each dataset, connectivity matrices are first computed at the segment level and then averaged within each subject. The subject-level matrices are subsequently averaged within each diagnostic group to obtain class-level connectivity patterns. This avoids biasing the visualization toward subjects with a larger number of segments.

% Fig.~\ref{fig:fc_analysis} shows representative class-averaged connectivity matrices for different frequency bands. The observed patterns vary across diagnostic groups, datasets, and frequency bands, suggesting that a single connectivity view may not capture all disease-relevant information.  This supports the use of a multi-view graph-bank design, where different connectivity metrics, frequency bands, and topology filters can contribute complementary information to the final subject-level prediction.

\subsection{Experimental Setup}

To evaluate the generalization of the proposed model, we use subject-wise 5-fold cross-validation for the AHEAP dataset. For CAUEEG, we follow the official non-overlap train/validation/test split from the original study, where subjects in the validation and test sets do not overlap with the training set. %The CAUEEG test set contains 90 subjects, including 35 HC, 33 MCI, and 22 dementia subjects. 
%we follow the official train/validation/test split from the original study and use the non-overlap version. Under these settings, segments from the same subject do not appear in both training and evaluation sets. 
All models are evaluated using the same subject-wise splits and three random seeds, and the final results are reported at the subject level. Models are trained using the Adam optimizer with early stopping based on validation loss.

% For a fair comparison, all models follow the same training protocol, subject-wise splits, and random seeds.
%All experiments were implemented in PyTorch and trained on an NVIDIA RTX 6000 Ada Generation GPU with CUDA acceleration.

\subsection{Node-feature and Graph-view Settings}
\textbf{Node-feature settings:} In this experiments, we use two groups of node features: relative band power (RBP), and statistical descriptors. The RBP features are computed from the power spectral density of each EEG channel using Welch's method~\cite{welch1967use}. 
We extract five frequency-band features: delta $\delta$ ($1$--$4$ Hz), theta $\theta$ ($4$--$8$ Hz), alpha $\alpha$ ($8$--$13$ Hz), beta $\beta$ ($13$--$30$ Hz), and gamma $\gamma$ ($30$--$45$ Hz).
The statistical descriptors summarize the amplitude distribution of each channel and include mean, standard deviation, skewness, kurtosis, minimum, maximum, and peak-to-peak amplitude.

\textbf{Graph-view settings:} The graph-view set $\mathcal{V}$ can be defined using different combinations of connectivity metrics, frequency bands, and topology filters. In this experiments, we use four graph views:
$$
\begin{aligned}
\mathcal{V}=
\{ & 
v_1 : (\mathrm{wPLI},\theta,\mathrm{complete}), v_2: (\mathrm{wPLI},\alpha,\mathrm{domain}),\\
& v_3: (\mathrm{coh},\alpha,\mathrm{hybrid}), v_4: (\mathrm{coh},\theta,\mathrm{top}\text{-}k)
\}
\end{aligned}
$$
These views were selected to provide a compact representation across different connectivity metrics, frequency bands, and topology filters. Coherence and wPLI are commonly used EEG functional-connectivity measures, and theta- and alpha-band connectivity have been frequently reported as informative for dementia-related EEG analysis~\cite{jiang_classification_2025,zheng_time-frequency_2025,paitel_functional_2025,mlinaric_eeg-based_2026}. The selected views are therefore used as candidate views for evaluating the proposed multi-view learning framework, rather than as the only possible graph-view configuration.

\subsection{Baseline and Comparison Models}
We compare the proposed framework with several raw-signal, feature-based, FC-based, and graph-based baselines.

\vspace{0.1cm}\noindent\textit{Classical machine learning and deep learning baselines:} Raw-signal baselines include a CEEDNet-style 1D-ResNet \cite{kim_deep_2023} and a CNN-LSTM. 
Feature-based baseline include 
%Random Forest and 
MLP trained on flattened channel-level node features, allowing us to assess performance without functional-connectivity information. 
We also include an ensemble baseline that combines CNN-LSTM with node-feature MLP to evaluate whether temporal EEG patterns and channel-level descriptors provide complementary information.
%\vspace{0.1cm}\noindent\textit{Deep learning baselines:} 
%\ann{For FC-based comparison, we train a CNN on the FC matrix as an image-like representation, motivated by prior studies using EEG functional-connectivity or connectome representations with convolutional models~\cite{jiang_classification_2025}. }

\vspace{0.1cm}\noindent\textit{Graph learning baselines:} We evaluate single-view GATv2 \cite{brody2021attentive} and LINKX-style models \cite{lim2021large} using one FC view at a time from the four-view candidate set $\mathcal{V}$. For each baseline, the model is trained separately on each candidate view, and the best subject-level performance among the four runs is reported. This ensures that the single-view baselines are compared using their strongest candidate view. For the LINKX baseline, we use the same no-skip LINKX-style encoder as in \method{}, so the comparison between LINKX and \method{} focuses on the effect of multi-view connectivity fusion.

\vspace{0.1cm}\noindent\textit{Multi-view learning baselines:} {We train two multi-view learning baselines. CNN is trained by stacking multiple view FC matrices as an image-like representation, and employs convolutional filters to learn features across multiple views for segment classification. This is motivated by previous studies using EEG functional-connectivity or connectome representations with convolutional models~\cite{jiang_classification_2025} using multi-view FC matrices. 
%CNN considers multi-view FC inputs as images, and employs convolutional filters to learn features across multiple views for segment classification}. 
The multi-view GATv2~\cite{brody2021attentive} model represents each EEG segment as multiple graphs with shared node features and different connectivity views. Each view is processed by a shared GATv2 encoder, and the resulting embeddings are fused for classification. %Multi-view GATv2 model is included to compare the proposed \method{} framework with a message-passing-based multi-view graph baseline. In this model, each EEG segment is represented by multiple graphs that share the same node features but use different connectivity views. Each view is processed by a shared GATv2 encoder, and the resulting view embeddings are fused for classification.
These baselines allow us to compare \method{} against alternative ways of integrating the same multi-view connectivity information.

\subsection{Performance Metrics}
% \hill{explain performance metrics used in the paper. We do not need to show formulas, but briefly expain what are they, and whey are they used here in the experiments.}
All reported results are evaluated at the subject level because diagnostic labels are defined for subjects rather than individual EEG segments. To account for imbalanced class distributions across diagnostic groups, we report balanced accuracy, which measures average recall across classes, and macro-F1, which averages class-wise F1-scores combining precision and recall.
%We report balanced accuracy and macro-F1 to account for imbalanced class distributions across diagnostic groups in both datasets.

\subsection{Results and Analysis}
\begin{table*}[]
\centering
\caption{Subject-level performance of \method{} and the baseline models, reported as mean $\pm$ standard deviation over three random seeds. Best and second-best results are shown in \textbf{bold} and \underline{\textit{underlined italics}}, respectively. For methods involving different Functional connectivity (``FC Input''), $\mathcal{V}$ denotes all four candidate graph views and $v^\star$ denotes the best-performing view among the four candidate graph views $\mathcal{V}$.
%For single-view LINKX and GATv2, $v^\star$ denotes the best-performing view among the four candidate graph views in $\mathcal{V}$.
}
% \caption{Subject-level prediction performance on CAUEEG and AHEAP. Values are reported as mean $\pm$ standard deviation over three random seeds using the specified segment window for each dataset. The best result in each metric column is shown in \textbf{bold}, and the second-best result is shown in \underline{\textit{underlined italics}}.}
\label{tab:mainresult}
\begin{tabular}{lccccccc}
\toprule
 &  & &  & \multicolumn{2}{c}{\textbf{CAUEEG (10-second window)}} & \multicolumn{2}{c}{\textbf{AHEAP (4-second window)}} \\
 &  & &  & \multicolumn{2}{c}{35 HC, 33 MCI, 22 Dementia} & \multicolumn{2}{c}{29 HC, 23 FTD, 36 AD} \\

\textit{Model} & \textit{RawEEG} & \textit{Node features} & \textit{FC input} & \multicolumn{1}{l}{\textit{Balanced Acc.}} & \multicolumn{1}{l}{\textit{Macro-F1}} & \multicolumn{1}{l}{\textit{Balanced Acc.}} & \multicolumn{1}{l}{\textit{Macro-F1}} \\ \midrule

1D ResNet & \checkmark & & & \underline{\textit{0.6146 $\pm$ 0.0135}} & \underline{\textit{0.6224 $\pm$ 0.0141}} & 0.5221 $\pm$ 0.0432 & 0.4815 $\pm$ 0.0511 \\

CNN-LSTM & \checkmark & & & 0.5700 $\pm$ 0.0133 & 0.5585 $\pm$ 0.0049 & 0.5697 $\pm$ 0.0527 & 0.5494 $\pm$ 0.0438 \\

% Random Forest & & \checkmark & & 0.5434 $\pm$ 0.0077 & 0.5558 $\pm$ 0.0085 & 0.5416 $\pm$ 0.0125 & 0.5065 $\pm$ 0.0211 \\

MLP-node & & \checkmark & & 0.5949 $\pm$ 0.0233 & 0.5961 $\pm$ 0.0199 & 0.5740 $\pm$ 0.0477 & 0.5550 $\pm$ 0.0446 \\

Ensemble & \checkmark & \checkmark & & 0.6084 $\pm$ 0.0285 & 0.6048 $\pm$ 0.0303 & 0.5791 $\pm$ 0.0258 & 0.5528 $\pm$ 0.0273 \\

CNN & & & $\mathcal{V}$ \ & 0.5210 $\pm$ 0.0243	
 & 0.5100 $\pm$ 0.0241 & 0.4758 $\pm$ 0.0312	
 & 0.4596 $\pm$ 0.0429 \\
 
LINKX & & \checkmark & $v^\star$& 0.6077 $\pm$ 0.0198 & 0.6154 $\pm$ 0.0179 %0.5747 $\pm$ 0.0129 & 0.5777 $\pm$ 0.0131 
% & {{0.5909 $\pm$ 0.0683}} & {{0.5723 $\pm$ 0.0653}} \\
&\underline{\textit{0.6060 $\pm$ 0.0460}} & 0.5806 $\pm$ 0.0384 \\

GATv2 & & \checkmark & $v^\star$ & 0.5910 $\pm$ 0.0236 & 0.5868 $\pm$ 0.0190 & 0.5383 $\pm$ 0.0255 & 0.4901 $\pm$ 0.0393 \\

Multi-view GATv2 & & \checkmark &  $\mathcal{V}$ & 0.5916 $\pm$ 0.0320 & 0.5992 $\pm$ 0.0316 & {0.6041 $\pm$ 0.0303} &  \underline{\textit{0.5895 $\pm$ 0.0187}} \\
\method{} (proposed) & & \checkmark &  $\mathcal{V}$ & \textbf{0.6649 $\pm$ 0.0294} &\textbf{ 0.6665 $\pm$ 0.0389} 
% & 0.5910 $\pm$ 0.0511 & 0.5701 $\pm$ 0.0475 
& \textbf{0.6191 $\pm$ 0.0281}& \textbf{0.5990 $\pm$ 0.0257} %(temp=2.5)

% Multi-view GATv2 & RBP+Stat & \begin{tabular}[c]{@{}l@{}}wpli\_theta\_full\\ wpli\_alpha\_fixed\\ coherence\_alpha\_combined\\ coherence\_theta\_topk4\end{tabular} & 0.5916 ± 0.0320 & 0.5992 ± 0.0316 & \multicolumn{1}{l}{} & \multicolumn{1}{l}{} \\
% M-LINKX (proposed) & RBP+Stat & \begin{tabular}[c]{@{}l@{}}wpli\_theta\_full\\ wpli\_alpha\_fixed\\ coherence\_alpha\_combined\\ coherence\_theta\_topk4\end{tabular} & 0.6649 ± 0.0294 & 0.6665 ± 0.0389 & 0.5910 ± 0.0511 & 0.5701 ± 0.0475 
\\
\bottomrule
\end{tabular}
\end{table*}

\subsubsection{Balanced Accuracy and Macro-F1 Scores}
Table~\ref{tab:mainresult} reports the subject-level classification performance on CAUEEG and AHEAP.
% The default segment durations are 10 seconds for CAUEEG and 4 seconds for AHEAP, respectively, so that each segment contains 2000 time points in both datasets. 
As shown in the table, \method{} achieves the best balanced accuracy and macro-F1 on both datasets. 

For CAUEEG, \method{} obtains balanced accuracy of $66.49\%$ and macro-F1 of $66.65\%$, outperforming the strongest non-proposed baseline by about $5.0\%$ in balanced accuracy and $4.4\%$ in macro-F1. In AHEAP, \method{} also obtains the highest balanced accuracy and macro-F1, although the improvement over the strongest graph-based baselines is smaller. This suggests that the proposed multi-view LINKX-style design is effective across both datasets, but the magnitude of improvement is dataset-dependent.

Several patterns can be observed from Table~\ref{tab:mainresult}. First, under the main experimental settings, \method{} improves over the node-feature-only MLP, suggesting that functional connectivity can provide complementary information beyond channel-level features when integrated effectively. Second, \method{} outperforms single-view LINKX and single-view GATv2, indicating that one connectivity view may not fully capture heterogeneous FC patterns across metrics, frequency bands, and topology filters. Third, \method{} also improves over multi-view GATv2, suggesting that LINKX-style separate encoding of node features and adjacency-based connectivity information is effective for the proposed multi-view FC representation.

% This suggests that functional connectivity provides useful relational information that complements local channel-level EEG features. 

%\ann{On AHEAP, coherence-theta top-k consistently receives relatively high weight, whereas the wPLI-based views receive lower but nonzero weights.}

\subsubsection{Subject Classification Confusion Matrices}
\begin{figure}
    \centering
    \includegraphics[width=0.95\linewidth]{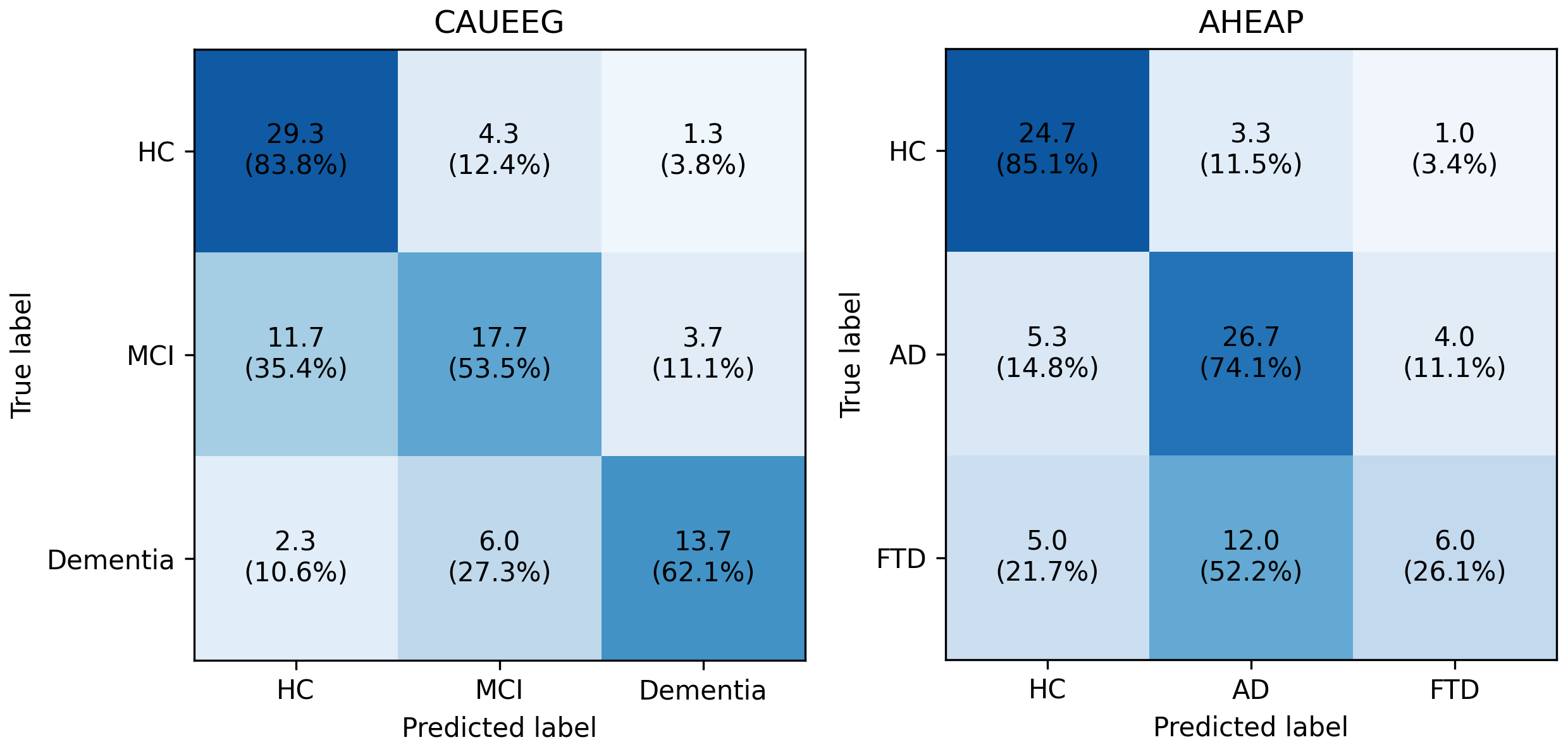}
    \caption{Subject-level confusion matrices of \method{} on CAUEEG (10$s$) and AHEAP (4$s$). Cells show average subject counts and row-normalized percentages over three random seeds. For AHEAP, confusion matrices are first summed across folds within each seed.}
    \label{fig:confmat}
\end{figure}

Fig.~\ref{fig:confmat} reports the normalized subject-level confusion matrices of \method{}. 
On CAUEEG, most errors involve the MCI class. In particular, a noticeable portion of MCI subjects is predicted as HC, and some dementia subjects are predicted as MCI. This suggests that MCI remains difficult to distinguish because it may share EEG characteristics with both healthy aging and dementia.
Similarly, in AHEAP, the model performs well for HC and AD, but FTD shows substantially lower recall. Most FTD errors are predicted as AD, indicating that the current connectivity metrics, frequency bands, and topology filters may capture general dementia-related abnormalities but are less discriminative for separating dementia subtypes. This indicates that additional features or adaptive graph-view selection may be needed for more fine-grained AD--FTD discrimination.

\subsubsection{Algorithm Performance \textit{vs.} Segment Lengths}
Although \method{} performs best under the main settings, Fig.~\ref{fig:duration-short} shows that this advantage is not consistent across all segment durations.
% Fig.~\ref{fig:duration-short} further analyzes the effect of segment duration on the proposed \method{}. 
Overall, \method{} remains competitive across different durations on both datasets, but its advantage over the baselines becomes less consistent as the window length increases. This suggests that increasing the segment length may stabilize some feature-based baselines, but does not necessarily improve multi-view connectivity learning.

On CAUEEG, \method{} achieves its best performance with the 10-second window. As the duration increases, its performance decreases, while several baselines remain relatively stable. This may indicate that longer windows provide more stable channel-level feature estimates, but may also smooth or average out transient connectivity patterns that are useful for the proposed multi-view graph representation.

For AHEAP, some baseline models improve as the segment duration increases, whereas \method{} shows a decreasing trend. One possible explanation is that longer windows stabilize node-feature or single-view representations, while multi-view FC graphs may become less distinctive when dynamic connectivity patterns are averaged over a longer interval.

% \subsubsection{Multi View Weights}
\subsection{Learned Graph-view Weights}

Fig.~\ref{fig:weight-short} shows the learned graph-view weights of \method{} across segment durations. The weights are not uniform, indicating that different connectivity views contribute differently to the final prediction. This supports multi-view connectivity modeling, as different metric-band-topology views may contribute differently across datasets and segment lengths.

On CAUEEG, the weight distribution changes noticeably across durations. For the 10-second setting, \method\ performs best and the coherence-alpha hybrid view receives the largest weight. When the duration increases to 20 seconds, this dominant weight drops and the view weights become more balanced. This may partly explain the performance decrease at longer durations. Prior work on dynamic functional connectivity has shown that window length strongly affects connectivity estimates and that wider windows can reduce temporal resolution~\cite{shakil2016evaluation}. Therefore, longer EEG segments may average over transient connectivity patterns and make graph views less distinctive.

The learned weights for AHEAP are more evenly distributed. The highest-weighted view also changes across durations: wPLI-alpha domain is largest at 2 seconds, while coherence-theta top-$k$ is largest at 4 and 6 seconds. This suggests that the most useful connectivity view can change with segment duration. These results also motivate future adaptive view selection instead of using one global view-weight vector for all subjects and disease groups.

\begin{figure}
    \centering
    \includegraphics[width=\linewidth]{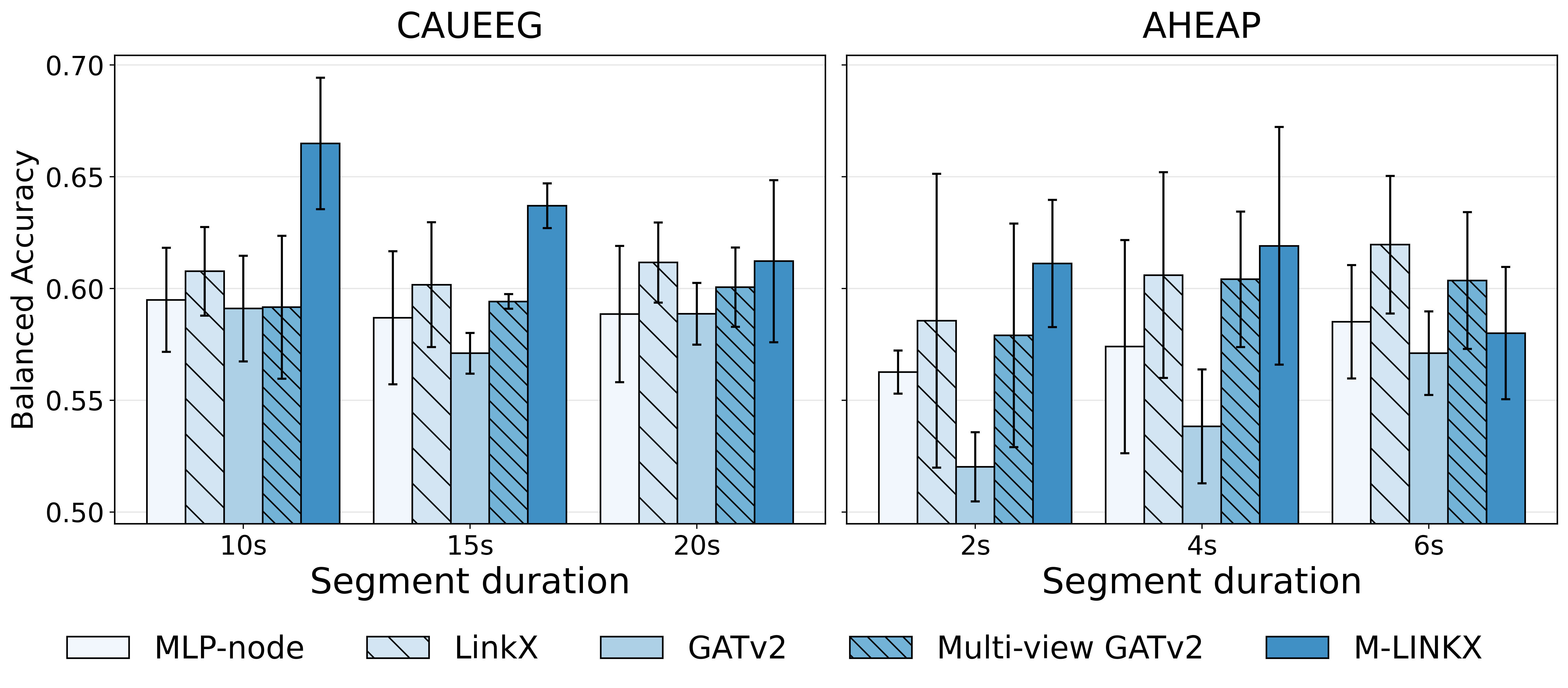}
    \caption{Subject-level balanced accuracy comparisons between \method\ and baseline models \textit{w.r.t.} different segment durations on CAUEEG and AHEAP. $y$-axis denotes mean balanced accuracies over three random seeds, and error bars denote standard deviation.}
    \label{fig:duration-short}
\end{figure}

\begin{figure}
    \centering
    \includegraphics[width=\linewidth]{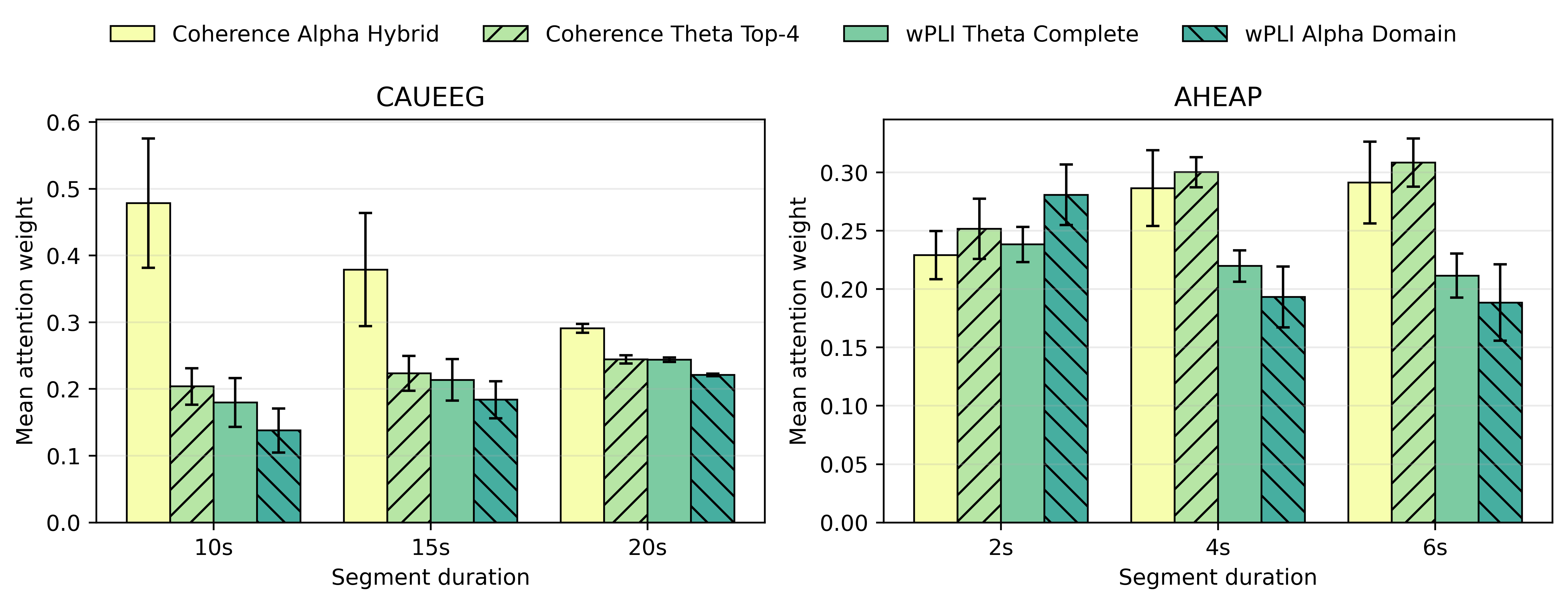}
    \caption{\method{} trainable graph-view weights across segment durations. Bars denote the mean weight over three random seeds, and error bars denote standard deviation. 
    }
    \label{fig:weight-short}
\end{figure}

% \begin{figure}
%     \centering
%     \includegraphics[width=\linewidth]{new_duration_bar_Balance_Acc.png}
%     \caption{Subject-level performance across segment durations on CAUEEG (left) and AHEAP (right). Bars indicate the mean over three random seeds, and error bars indicate the standard deviation.}
%     \label{fig:duration}
% \end{figure}
% \begin{figure}
%     \centering
%     \includegraphics[width=\linewidth]{candidate_attention_new.png}
%     % \includegraphics[width=\linewidth]{candidate_attention.png}
%     \caption{\method{} trainable graph-view weights across segment durations on CAUEGG and AHEAP. Bars denote the mean weight over three random seeds, and error bars denote standard deviation. 
%     }
%     \label{fig:weight}
% \end{figure}

% % \subsection{Ablation Study}
% \ann{performance of (i) each single view in the multi-view (linkx) ; (ii) difference banks of views: -metric-band-topology}
% \ann{-compare: 1 segment = 1 graph vs. 1 subject = 1 graph??}
% \textcolor{red}{Figure - Proposed model vs Different banks vs single candidate linkx}

\subsection{Discussion}
% What do these results mean, and why might they happen?
% Main interpretation: M-LINKX works well because it combines node features and multiple FC views, but the benefit depends on how FC is modeled.
% Connectivity is useful but not universally better: Some duration settings show MLP-node can be competitive or better than graph models. So FC is helpful only when graph construction and architecture are appropriate.
% Duration matters: FC values and retained topology can change with segment length. Therefore, window length affects connectivity quality and model performance.
% View weights suggest heterogeneous contribution: Different metrics, bands, and topology filters contribute differently, but learned weights should be interpreted as model behavior, not direct biological proof.
% Remaining difficulty: HC/MCI and AD/FTD are hard to separate, so future features may need to be more subtype-sensitive.

The results suggest that multi-view functional-connectivity learning can improve EEG segment-based dementia classification when the connectivity information is modeled appropriately. Meanwhile, the learned view weights are relatively stable across segment lengths and are not uniform, indicating that \method{} uses the graph views unequally during fusion. These weights reflect the behavior of the model and should not be interpreted as direct biological importance.

The duration analysis further shows that the contribution of connectivity-based models varies across segment lengths. Since functional-connectivity estimates and retained graph topologies can change with the EEG window duration, the optimal segment length may depend on the dataset, disease groups, and model architecture. The confusion matrices also show that some diagnostic groups remain difficult to separate, especially clinically related classes such as HC/MCI and AD/FTD.
\section{Conclusion and Future Work}
% What is the final takeaway of the paper, and what is next?
% What proposed: M-LINKX.
% What it does: multi-view FC graph learning for EEG dementia classification.
% What is found: best performance under main settings on two datasets.
% Main takeaway: multi-view FC is useful when integrated properly, but its benefit depends on duration/model design.
% Future work

In this paper, we proposed \method{}, a multi-view graph learning framework for EEG-based dementia classification. We argue that EEG data are noisy and the correlation between channels (\textit{i.e.}, electrodes) may provide useful information to benefit subject classification. Instead of finding a single best functional connectivity view, \method\ represents each EEG segment using channel-level node features and multiple functional-connectivity graph views constructed from different connectivity metrics, frequency bands, and topology filters. By leveraging the multi-view graph topologies and a simple heterophilic graph learning backbone, which treated node features and adjacency matrices as two separated information sources, \method\ adaptively learns and aggregates features from different views, and further concatenates node features for segment level prediction.  
Experiments on two dementia-related EEG datasets show that %\method{} achieves the best subject-level performance under the main experimental settings. These results suggest that 
multi-view functional connectivity can improve EEG-based classification when it is integrated with appropriate graph-learning architectures. Our analysis also suggests that segment lengths and edge construction strategies play important roles for graph based methods.
%the benefit of connectivity-based modeling depends on the segment length and graph construction strategy.

Future work will explore adaptive graph-view selection, where the model can assign different importance to connectivity views across datasets, subjects, or disease groups. %We also plan to investigate additional EEG representations that may better distinguish clinically related groups such as HC/MCI and AD/FTD.

\section*{Data Availability}
The datasets used in our study are publicly accessible from the following sources:
\begin{itemize}[leftmargin=*, labelsep=0.5em, itemsep=0pt, topsep=2pt]
\item {AHEAP}: \href{https://openneuro.org/datasets/ds004504/versions/1.0.8}{OpenNeuro ds004504, v1.0.8}. 
\item {CAUEEG}: Available from the authors upon request~\cite{kim_deep_2023}.
\end{itemize}

\section*{Acknowledgment}
This work has been supported in part by the U.S. National Science Foundation (NSF) under Grant Nos.\, IIS-2236579, IIS-2302786, and IOS-2430224.
% The AHEAP dataset is publicly available on OpenNeuro at
% \url{https://openneuro.org/datasets/ds004504/versions/1.0.8}. 
% The CAUEEG dataset is available upon reasonable request from the authors of the original dataset publication \cite{kim_deep_2023}. 

\begingroup
\footnotesize
\bibliographystyle{IEEEtran}
\bibliography{ref}
\endgroup
\end{document}